\documentclass[11pt]{article}

\usepackage[final]{acl}

\usepackage{times}
\usepackage{latexsym}

\usepackage[T1]{fontenc}

\usepackage[utf8]{inputenc}
\usepackage{CJKutf8}
\usepackage{enumitem}

\usepackage{microtype}

\usepackage{inconsolata}

\usepackage{graphicx}

\usepackage{tcolorbox}
\usepackage{caption}
\tcbuselibrary{skins,breakable}

\usepackage{pifont}
\usepackage{amsmath}
\usepackage{amssymb}  
\usepackage{bm}       
\usepackage{mathtools} 
\usepackage{xspace}
\usepackage{cleveref}
\usepackage{booktabs}
\usepackage{multirow}
\usepackage{graphicx}
\usepackage{tablefootnote}
\usepackage[normalem]{ulem}
\useunder{\uline}{\ul}{}

\crefformat{section}{\S#2#1#3} 
\crefformat{subsection}{\S#2#1#3}
\crefformat{subsubsection}{\S#2#1#3}
\usepackage{scalerel,xparse}

\NewDocumentCommand\emoji{}{%
  \raisebox{-0.2\height}{\includegraphics[width=1.2em]{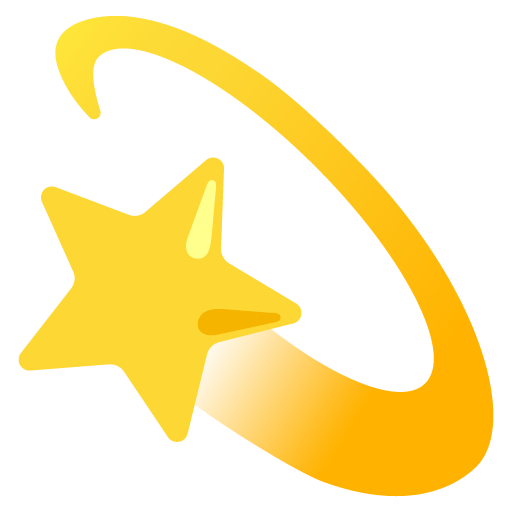}}%
}

\NewDocumentCommand\thinkingemoji{}{%
  \raisebox{-0.2\height}{\includegraphics[width=1em]{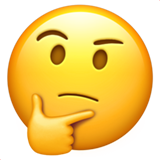}}%
}
\newcommand{\symboltwo}{\textsuperscript{\textdagger}}

\definecolor{oxfordblue}{rgb}{0.0, 0.13, 0.28}
\definecolor{harvardcrimson}{rgb}{0.79, 0.0, 0.09}
\definecolor{dartmouthgreen}{rgb}{0.05, 0.5, 0.06}
\definecolor{princetonorange}{rgb}{1.0, 0.56, 0.0}
\definecolor{yaleblue}{rgb}{0.06, 0.3, 0.57}
\definecolor{usccardinal}{rgb}{0.6, 0.0, 0.0}
\definecolor{uclablue}{rgb}{0.33, 0.41, 0.58}
\definecolor{msugreen}{rgb}{0.09, 0.27, 0.23}
\definecolor{cornellred}{rgb}{0.7, 0.11, 0.11}
\definecolor{pomegranate}{RGB}{192, 57, 43}
\definecolor{anti-pomegranate}{RGB}{43,178,192}
\definecolor{alizarin}{RGB}{231, 76, 60}
\definecolor{anti-belize}{RGB}{185, 41, 56}
\definecolor{belize}{RGB}{41, 128, 185}
\definecolor{peter}{RGB}{52, 152, 219}
\definecolor{green}{RGB}{22, 160, 133}
\definecolor{anti-green}{RGB}{160,22,118}
\definecolor{turquoise}{RGB}{26, 188, 156}
\definecolor{pumpkin}{RGB}{211, 84, 0}
\definecolor{anti-pumpkin}{RGB}{0,22,211}
\definecolor{carrot}{RGB}{230, 126, 34}
\definecolor{wisteria}{RGB}{142, 68, 173}
\definecolor{anti-wisteria}{RGB}{99,173,68}
\definecolor{amethyst}{RGB}{155, 89, 182}
\definecolor{nephritis}{RGB}{39, 174, 96}
\definecolor{anti-nephritis}{RGB}{174,39,117}

\newcommand{\ywj}[1]{{\color{black} #1}}

\title{\emoji SSR-Zero: Simple Self-Rewarding Reinforcement Learning for Machine Translation
}

\author{Wenjie Yang, Mao Zheng, Mingyang Song, Zheng Li, Sitong Wang\symboltwo \\
  Tencent Hunyuan, 
  Columbia University\symboltwo \\
  \texttt{leonzxyang@tencent.com} \\}

\begin{document}
\maketitle
\begin{abstract}
Large language models (LLMs) have recently demonstrated remarkable capabilities in machine translation (MT). However, most advanced MT-specific LLMs rely heavily on external supervision during training, such as human-annotated reference data or trained reward models (RMs), which are expensive to obtain and difficult to scale.
To address this limitation, we propose \underline{S}imple \underline{S}elf-\underline{R}ewarding (\emoji \textbf{SSR}), a reinforcement learning (RL) framework for MT that is reference-free and relies solely on self-judging rewards. Using only 13K monolingual examples and Qwen-2.5-7B as the backbone, SSR-Zero-7B outperforms existing MT-specific LLMs as well as larger general LLMs such as Qwen2.5-32B-Instruct on English $\leftrightarrow$ Chinese translation benchmarks including WMT23, WMT24, and FLORES200. It further demonstrates strong generalization to low-resource language pairs.
In addition, when augmented with external supervision from COMET, our strongest model, SSR-X-Zero-7B, surpasses all existing open-source models under 72B parameters and performs competitively with leading closed-source systems in English $\leftrightarrow$ Chinese translation. Our analysis highlights the effectiveness and generalizability of the self-rewarding mechanism relative to external LLM-as-a-judge approaches and demonstrates its complementary benefits when combined with trained RMs. We will publicly release our code, data, and models.

\end{abstract}

\section{Introduction}

Large language models (LLMs) have recently achieved substantial progress in machine translation (MT) \cite{aryabumi2024aya,rei2024tower,cui2025multilingual}, benefiting from large-scale pre-training and effective transfer of multilingual knowledge. MT-specific LLMs such as Tower and X-ALMA further improve translation quality through continual pre-training (CPT) on billions of parallel and monolingual tokens, followed by fine-tuning on high-quality human-annotated data \cite{alves2024tower,cui2025multilingual}. \ywj{While effective, this paradigm relies heavily on parallel data, which -- even when available at scale -- often suffers from noise and semantic misalignment \cite{meng2024noisy}, machine-generated contamination \cite{thompson2024shocking}, and translationese artifacts \cite{koppel2011translationese}, limiting its reliability as a supervision signal \cite{uhlig2025crosslingual}.}

In parallel, recent advances in inference-time reasoning have shown that reinforcement learning (RL) can substantially enhance LLM capabilities. Models such as OpenAI o1 \cite{jaech2024openai} and DeepSeek R1 \cite{guo2025deepseek} employ R1-style training with RL algorithms (e.g., GRPO \cite{shao2024deepseekmath}, DAPO \cite{yu2025dapo}) to incentivize reasoning behaviors, achieving strong performance in tasks such as logic, coding, and mathematics \cite{guo2025deepseek,xie2025logic,song2025fastcurl}. Recent work has begun extending these ideas to MT, either by introducing explicit reasoning patterns \cite{wang2024drt,wang2025deep} or by allowing models to learn reasoning implicitly during training \cite{feng2025mt}. However, existing RL-based MT approaches still depend heavily on external supervision, either in the form of human-annotated references or trained reward models distilled from expensive labeled data, which limits their scalability.


To address this limitation, we propose Simple Self-Rewarding (\textbf{SSR}), a RL framework for MT that eliminates the need for any external supervision. SSR adopts a self-judging mechanism in which the LLM itself evaluates its translation outputs and produces reward signals, which are then used to optimize the model via GRPO. Using only 13K monolingual sentences (6.5K English and 6.5K Chinese), we train uninstructed Qwen2.5-7B and 3B models, resulting in SSR-Zero-7B and SSR-Zero-3B. SSR-Zero-7B improves its backbone by 18.11\% on Chinese-to-English and 14.74\% on English-to-Chinese translation.

%

Extensive experiments on WMT23, WMT24, and FLORES-200 show that SSR-Zero-7B outperforms existing MT-specific LLMs such as TowerInstruct-13B and GemmaX-28-9B, as well as larger general-purpose LLMs including Qwen-2.5-32B-Instruct. When augmented with external COMET rewards, our strongest model, SSR-X-Zero-7B, achieves the best performance among evaluated open-source LLMs under 72B parameters for English $\leftrightarrow$ Chinese translation and performs competitively with closed-source systems such as GPT-4o and Gemini 1.5 Pro. Results on SSR-Zero-3B further indicate that SSR generalizes to smaller base models with weaker judging capabilities.

%
Finally, we conduct detailed analyses to examine the effectiveness and generalizability of self-rewarding. These include experiments on low-resource languages (Gujarati and Kazakh), comparisons between self-rewarding and external reward models, and an investigation into the impact of reference-based versus referenceless rewards.


In summary, \textbf{our contributions are}:
1) We propose SSR, a self-rewarding RL framework for MT that removes reliance on external reward models and reference translations.
2) We demonstrate that SSR substantially improves MT quality across model sizes, language pairs, and resource settings, outperforming strong open-source MT baselines.
3) We show that self-generated rewards complement external rewards, enabling SSR-X-Zero-7B to achieve the strongest performance among evaluated open-source models for English $\leftrightarrow$ Chinese translation.
4) We provide a systematic analysis of reward design choices for RL-based MT and release our code, data, and models to support future research.

\begin{figure*}[t]
  \centering
  \includegraphics[width=0.9\textwidth]{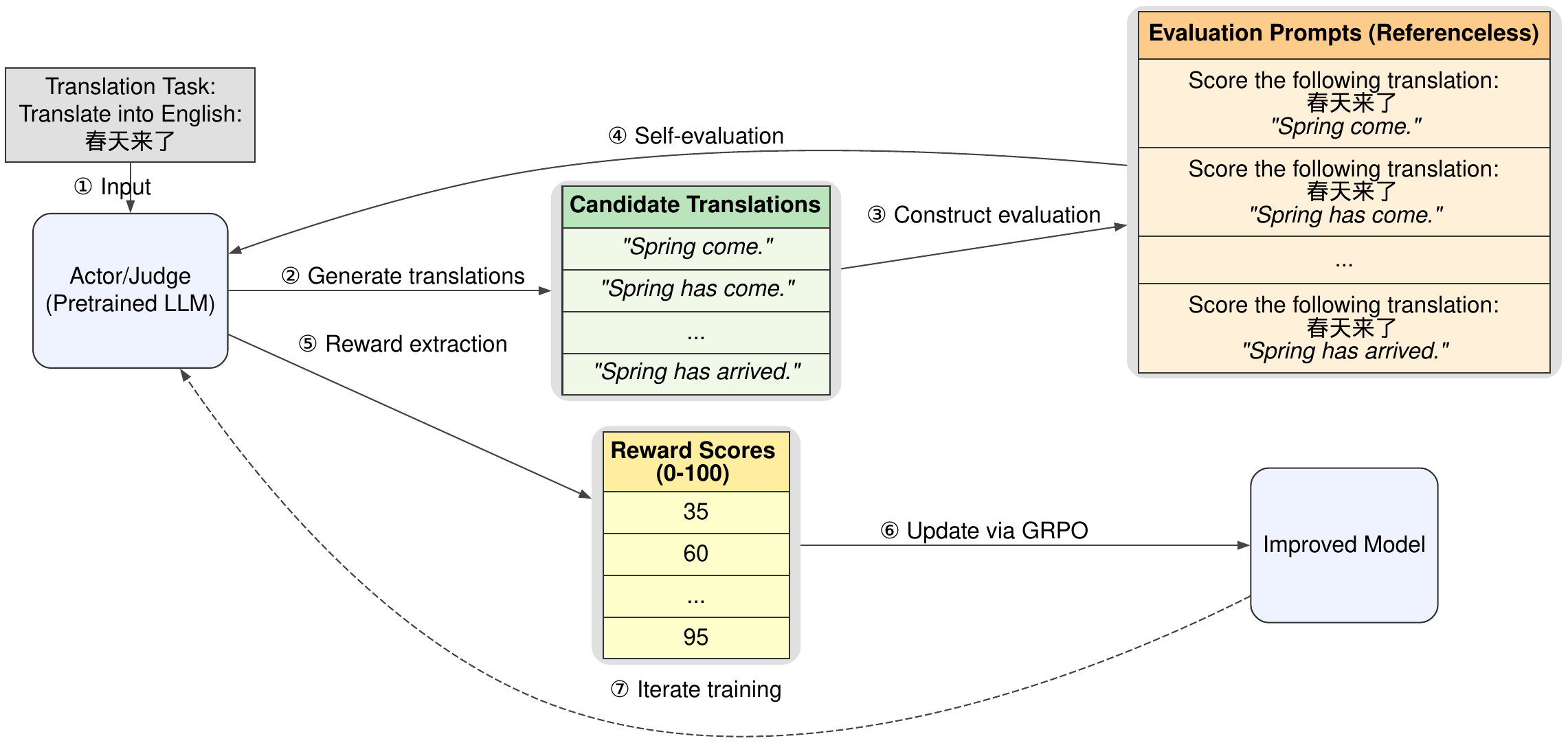}
  \caption{Overview of the \emoji SSR framework. SSR is an R1-Zero-like RL training method for machine translation, which uses the same model as both actor and judge. It does not require external reward models or human-annotated reference data. Prompts shown here are simplified for clarity.}
  \label{fig:intro-ssr-pipeline}
\end{figure*}

\section{Related Work}

\subsection{Machine Translation with LLMs}

Recent advances in large language models (LLMs) have substantially improved machine translation (MT) across many language pairs \cite{costa2022no, lu-etal-2024-llamax, workshop2022bloom}. Many strong MT-focused LLMs \cite{rei2024tower,cui2025multilingual} rely on continual pre-training (CPT) over large-scale mixtures of parallel and monolingual data, often exceeding tens of billions of tokens.

Beyond data scale, prior work has shown that enriching training objectives can further enhance MT performance. For example, \citet{rei-etal-2024-tower} incorporate auxiliary tasks such as translation evaluation, MQM-style error detection, and named entity recognition, while \citet{cui2025multilingual} propose a sequential data mixing strategy that prioritizes parallel data during CPT, yielding competitive performance with commercial systems such as Google Translate and GPT-4-turbo.

Despite their strong performance, these approaches depend heavily on large volumes of curated or annotated data. As training scales, the cost and availability of such resources increasingly limit the sustainability of MT model development.

%
%

\subsection{MT via Reinforcement Learning}

Reinforcement learning (RL) has long been explored in MT to mitigate exposure bias in supervised training \cite{bengio2015scheduled}. Early work applied algorithms such as REINFORCE \cite{ranzato2015sequence}, actor–critic methods \cite{bahdanau2016actor}, and policy gradients \cite{yu2017seqgan}, using either rule-based metrics (e.g., BLEU, ROUGE) or trained reward models \cite{wu2017sequence}.

More recently, in the context of LLMs, R1- and R1-Zero-style training has demonstrated that RL algorithms such as GRPO, combined with verifiable rewards, can substantially improve reasoning ability \cite{guo2025deepseek}. This paradigm has been extended to MT, either by introducing explicit reasoning patterns or by allowing models to learn latent reasoning processes during RL training.

For example, \citet{he2025r1} train MT models using manually designed chain-of-thought data with COMET-based rewards, while \citet{feng2025mt} explore BLEU-, COMETKiwi-, and hybrid reward signals, achieving strong results with MT-R1-Zero-Sem. \citet{wang2025deep} further propose using a large LLM-based judge to evaluate both reasoning steps and translation outputs during RL, targeting literary translation.

Nevertheless, existing RL-based MT methods still rely on external supervision signals, such as reference translations, trained reward models, or large frozen judges. \ywj{Beyond cost, available parallel data often suffers from noise \cite{meng2024noisy}, machine-generated contamination \cite{thompson2024shocking}, and translationese artifacts \cite{koppel2011translationese}, limiting its reliability as supervision. Recent work has also explored human-preference alignment for MT via RLHF \cite{ramos2024aligning,xu2024advancing} and direct quality optimization \cite{uhlig2025crosslingual}, but these approaches still depend on annotated preference data.}

\subsection{Self-Judging in RL}

Recent work has investigated self-rewarding or self-judging mechanisms, in which LLMs generate their own feedback signals for training \cite{chen2024self, wu2024meta, Zhang2025ProcessbasedSL}. Such approaches aim to reduce reliance on human annotations or reward models distilled from human judgments. For instance, \citet{chen2024self} iteratively perform self-instruction sampling, self-judging, and DPO training, demonstrating improvements in both instruction-following and evaluation capabilities.

Related self-improving paradigms, including self-play and self-judging, have shown effectiveness in domains such as mathematical reasoning \cite{zhang2025right, zhao2025absolute}, vision–language alignment \cite{zhou2024calibrated}, and cross-lingual transfer \cite{chen2024self, geng2024not, yang2024language}. However, self-judging remains relatively underexplored for MT.

One notable exception is \citet{zou2025trans}, who propose a self-play framework based on Monte Carlo Tree Search to derive preferences from cross-lingual semantic consistency. While effective, their approach does not outperform MT-specific LLMs such as TowerInstruct when using the same base model.

In contrast, our approach eliminates the need for external supervision, operates fully online, and achieves strong performance using only monolingual data. These results suggest that sufficiently strong pre-trained LLMs already possess usable translation and MT-evaluation capabilities, pointing toward a viable path for self-improving MT without human feedback.

\section{Methodology}

In this section, we first outline the SSR methodology (\cref{sec:method:pipeline}), followed by an introduction of the reward design within the RL framework (\cref{sec:method:reward}). Finally, we introduce the RL algorithm employed in our work (\cref{sec:method:grpo}).

\subsection{Simple Self-Rewarding (SSR)}\label{sec:method:pipeline}
SSR is a R1-Zero-like RL approach with a novel self-evaluation mechanism that simplifies reward signal acquisition. This mechanism leverages a pre-trained LLM that alternates between acting as both an actor and a judge.

As illustrated in Figure \ref{fig:intro-ssr-pipeline}, the pretained model, at each training step, first plays the role of an actor that accepts a batch of translation prompts (\ding{192}). For each prompt, the model generates a group of N candidate translations (\ding{193}). These candidate translations are then constructed on LLM-as-a-judge prompts separately (\ding{194}).
Next, the model switches to a judge role, evaluating all prompts to estimate translation quality and generate judgments (\ding{195}). Each judgment includes a score from 0 to 100, where 0 indicates poor translation and 100 indicates perfect translation. We extract reward scores from judgments using regular expressions (\ding{196}) and then use them in the RL algorithm (i.e., GRPO) to update the actor model's parameters (\ding{197}). In total, one translation prompt generates N candidate translations and N reward scores. We iterate Step \ding{192} through \ding{197} multiple times until the model's performance converges (\ding{198}).


Below are the prompts for generating translations (i.e., \textit{actor prompt}) and evaluations (\textit{judge prompt}) used in SSR training. The actor prompt builds on Deepseek-R1-Zero's system prompt \cite{guo2025deepseek}, requiring the model to answer within a specific format (i.e., \texttt{<answer></answer>}) and think before responding.

\begin{tcolorbox}[
	colframe=gray!80!black, 
	colback=gray!10!white, 
	coltitle=white, 
	fonttitle=\bfseries, 
	title=Actor Prompt: Generating Translations, 
	boxrule=0.5mm, 
	]
A conversation between User and Assistant. The User asks a question, and the Assistant solves it. 
The Assistant first thinks about the reasoning process in the mind and then provides the User with the answer. 
The reasoning process is enclosed within \texttt{<think> </think>} and answer is enclosed within \texttt{<answer> </answer>} tags, respectively, 
i.e., \texttt{<think>} reasoning process here \texttt{</think> <answer>} answer here \texttt{</answer>}. \\
\\
User:\\
Translate the following text to \{tgt\_lang\}: \\
\{src\_text\}
\\
Assistant:
\end{tcolorbox}
\begin{tcolorbox}[
    enhanced, 
    breakable,
    skin first=enhanced,
    skin middle=enhanced,
    skin last=enhanced,
    title={Judge Prompt: Self-Evaluating},
colframe=gray!80!black, 
	colback=gray!10!white, 
	coltitle=white, 
	fonttitle=\bfseries, 
	boxrule=0.5mm, 
]
A conversation between User and Assistant. The User asks a question, and the Assistant solves it. 
The Assistant first thinks about the reasoning process in the mind and then provides the User with the answer. 
The reasoning process is enclosed within \texttt{<think> </think>} and answer is enclosed within \texttt{<answer> </answer>} tags, respectively, 
i.e., \texttt{<think>} reasoning process here \texttt{</think> <answer>} answer here \texttt{</answer>}. \\
\\
User:\\
Score the following translation from \{src\_lang\} to \{tgt\_lang\} on a continuous scale from 0 to 100, where a score of zero means ``no meaning preserved'' and score of one hundred means ``perfect meaning and grammar''.\\
Additionally, give a score of zero if the translation 1) contains irrelevant content, such as interpretations of the translation, 2) does not match the target language, 3) contains multiple translations.\\\\
\{src\_lang\} source: \{src\_text\} \\
\{tgt\_lang\} translation: \{translated\_text\}
\\
Assistant:

\end{tcolorbox}
The judge prompt is modified from GEMBA-DA \cite{kocmi-federmann-2023-large}, a widely-used LLM-as-a-judge template for direct assessment of translation, which achieved SOTA performance in translation quality assessment using GPT-4. Compared to GEMBA-DA, our judge prompt includes an  ``think-before-answer'' system instruction. This addition explicitly encourages the model to take advantage of the reasoning capabilities acquired during RL training when evaluating translations.   Additionally, we instruct the judge to give a zero score for unwanted candidate translations containing irrelevant content or language misalignment. During training, only the content within \texttt{<answer></answer>} tags is extracted and incorporated into the judge's instructions.


\subsection{Reward Modeling}\label{sec:method:reward}

Our RL training utilizes two types of rewards: \textit{self-reward} and \textit{format reward}.

\paragraph{Self Reward}
This reward estimates the quality of the model's translation using the training model itself, denoted by:

\[
r_{\text{self}} = \frac{M_{\text{self}}(\text{src, trans})}{100}, r_{\text{self}}  \in [0,1]
\]

where $M_\text{self}$ is the model during the training. Using the judge prompt, the model takes both source text and model translation (without reference translations) and generates a judgment containing a score on a 100-point scale. All rewards are then linearly rescaled to $[0,1]$ before GRPO.


\paragraph{Format Reward}
This reward checks whether the model generation follows the format defined in the actor prompt:

\[
r_{\text{format}} = 
\begin{cases}
1, & \text{if format is correct} \\
0, & \text{if format is incorrect}
\end{cases}
\]

\paragraph{Overall Reward}
In training, we combine the two types of rewards  to train our SSR-Zero model:

\[
r_{\text{all}} = 
\begin{cases}
r_{\text{self}}+r_{\text{format}}, & \text{if } r_{\text{format}} \neq 0 \\
0, & \text{if } r_{\text{format}} = 0
\end{cases}
\]
In addition, we investigate integrating external reward signals to further enhance model performance. Our strongest model, SSR-X-Zero (\underline{SSR} with e\underline{X}ternal rewards), incorporates rewards computed by COMET, an automatic MT evaluation metric \cite{rei-etal-2022-comet} that scores translation quality using source sentences, machine-generated translations, and reference translations:


\[
r_{\text{all}}' = 
\begin{cases}
r_{\text{self}} + r_{\text{COMET}}+r_{\text{format}}, & \text{if } r_{\text{format}} \neq 0 \\
0, & \text{if } r_{\text{format}} = 0
\end{cases}
\]
$$ r_\text{COMET} = M_{\text{COMET}}(\text{src, trans, ref}), r_\text{COMET} \in [0, 1]
$$

\subsection{RL algorithm}\label{sec:method:grpo}

We follow the work of \citet{shao2024deepseekmath} and \citet{guo2025deepseek} by adopting the Group Related Policy Optimization (GRPO) algorithm for training, as it demonstrates stability and strong performance.  Specifically, for each given translation prompt $p$, the policy model $\pi_{\theta_\text{old}}$ first samples a group of candidate translations $G$ $\{o^i\}^G_{i=1}$. Then, using the same policy model, we perform the SSR procedure described earlier to obtain rewards $\{r^i_{\text{all}}\}^G_{i=1}$ for all candidate translations. Next, we compute the advantage for the $i$-th candidate translation by normalizing the group-level rewards:

$$
A_{i} = \frac{r^i_{\text{all}} - \text{mean}\left(\{r^i_{\text{all}}\}_{i=1}^G\right)}{\text{std}\left(\{r^i_{\text{all}}\}_{i=1}^G\right)}
$$
Using these advantages, GRPO optimizes the policy by maximizing the following objective:

\begin{align*}
J_{\text{GRPO}}(\theta) &= \mathbb{E}_{q \sim P(Q), \{o^i\}_{i=1}^G \sim \pi_{\theta_{\text{old}}}(O|p)}  \\
& \Bigg[ \frac{1}{G} \sum_{i=1}^G \min \Bigg( \frac{\pi_{\theta}(o^i | p)}{\pi_{\theta_{\text{old}}}(o^i | p)} A_i, \\
&\quad \text{clip}\left( \frac{\pi_{\theta}(o^i | p)}{\pi_{\theta_{\text{old}}}(o^i | p)}, 1 - \varepsilon, 1 + \varepsilon \right) A_i \\
&\quad - \beta D_{\text{KL}}(\pi_{\theta} \| \pi_{\text{ref}}) \Bigg) \Bigg]
\end{align*}

where $\varepsilon$ and $\beta$ are hyperparameters, $\pi_\text{ref}$ is the reference model, and $D_{\text{KL}}(\pi_{\theta} \| \pi_{\text{ref}})$ is the KL divergence between $\pi_{\theta}$ and $\pi_{\text{ref}}$.


\section{Experiments}

\begin{table*}[t]
\resizebox{\textwidth}{!}{%
\begin{tabular}{@{}llllllllllllllll@{}}
\toprule
 & \multicolumn{7}{c}{ZH$\rightarrow$EN} &  & \multicolumn{7}{c}{EN$\rightarrow$ZH} \\ \cmidrule(lr){2-8} \cmidrule(l){10-16} 
Models & \multicolumn{2}{c}{WMT23} & \multicolumn{2}{c}{WMT24} & \multicolumn{2}{c}{Flores200} & \multicolumn{1}{c}{\multirow{2}{*}{Avg.}} &  & \multicolumn{2}{c}{WMT23} & \multicolumn{2}{c}{WMT24} & \multicolumn{2}{c}{Flores200} & \multicolumn{1}{c}{\multirow{2}{*}{Avg.}} \\ \cmidrule(lr){2-7} \cmidrule(lr){10-15}
 & KIWI & XCM & KIWI & XCM & KIWI & XCM & \multicolumn{1}{c}{} &  & KIWI & XCM & KIWI & XCM & KIWI & XCM & \multicolumn{1}{c}{} \\ \midrule
\multicolumn{16}{c}{\textbf{Closed-Source LLMs}} \\
Claude-3.5-Sonnet & 81.61 & 93.06 & 81.06 & 90.54 & 89.41 & 97.68 & \textbf{\underline{88.89}} &  & \textbf{80.15} & 92.00 & 80.00 & 86.31 & 89.47 & 94.32 & {\underline{87.04}} \\
GPT-4o & 80.92 & 92.15 & 79.90 & 89.06 & 88.94 & 96.50 & 87.91 &  & 76.71 & 88.56 & 77.42 & 83.95 & 88.30 & 93.30 & 84.71 \\
Gemini-1.5-Pro & 80.71 & 92.44 & 79.02 & 88.90 & 88.15 & 97.32 & 87.76 &  & 79.80 & 91.95 & 79.54 & 87.11 & 89.30 & 94.54 & {\underline{ 87.04}} \\
\multicolumn{16}{c}{\textbf{Open-Source LLMs}} \\
\textbf{General Purpose LLMs} &  &  &  &  &  &  &  &  &  &  &  &  &  &  &  \\
Qwen3-32B\thinkingemoji & 79.74 & 90.79 & 79.20 & 88.47 & 87.68 & 95.75 & 86.94 &  & 76.94 & 89.75 & 76.96 & 84.10 & 87.45 & 92.18 & 84.56 \\
Qwen3-32B & 80.28 & 91.95 & 79.95 & 89.53 & 88.88 & 97.18 & 87.96 &  & 79.27 & 91.28 & 79.51 & 86.63 & 89.69 & 94.07 & {\textbf{86.74}} \\
Qwen3-8B\thinkingemoji & 78.30 & 89.03 & 77.99 & 86.94 & 85.82 & 93.89 & 85.33 &  & 74.94 & 88.22 & 75.39 & 82.25 & 86.08 & 91.02 & 82.98 \\
Qwen3-8B & 79.87 & 91.42 & 79.58 & 89.02 & 88.61 & 96.55 & 87.51 &  & 78.59 & 90.90 & 78.71 & 85.31 & 88.90 & 93.30 & 85.95 \\
Qwen2.5-72B-Instruct & 80.62 & 92.14 & 80.46 & 90.06 & 88.90 & 97.28 & {\textbf{88.24}} &  & 78.18 & 91.34 & 78.18 & 85.13 & 88.04 & 93.20 & 85.68 \\
Qwen2.5-32B-Instruct & 77.73 & 89.28 & 78.77 & 88.69 & 87.13 & 95.50 & 86.18 &  & 77.73 & 90.23 & 78.77 & 83.48 & 87.13 & 91.99 & 84.89 \\
Qwen2.5-3B-Instruct & 73.52 & 86.60 & 75.82 & 85.03 & 85.46 & 93.41 & 83.31 &  & 66.78 & 84.34 & 67.67 & 76.12 & 78.79 & 85.19 & 76.48 \\
Qwen2.5-7B-Instruct & 77.56 & 89.40 & 76.71 & 87.12 & 86.28 & 94.06 & 85.19 &  & 73.81 & 88.11 & 72.98 & 80.93 & 85.18 & 89.90 & 81.82 \\
QwQ-32B\thinkingemoji & 74.61 & 85.12 & 75.08 & 84.34 & 80.88 & 89.21 & 81.54 &  & 77.33 & 89.10 & 78.13 & 85.03 & 86.51 & 90.93 & 84.51 \\
Gemma2-27B-it & 80.32 & 91.96 & 79.42 & 89.14 & 88.64 & 96.72 & 87.70 &  & 76.95 & 90.50 & 77.38 & 84.17 & 87.79 & 92.51 & 84.88 \\
Gemma2-9B-it & 79.86 & 91.21 & 79.25 & 88.41 & 88.32 & 96.25 & 87.22 &  & 75.22 & 89.66 & 74.15 & 81.65 & 85.95 & 90.90 & 82.92 \\
 &  &  &  &  &  &  &  &  &  &  &  &  &  &  &  \\
\textbf{MT-Specific LLMs} &  &  &  &  &  &  &  &  &  &  &  &  &  &  &  \\
TowerInstruct-7B-v0.2 & 77.78 & 89.13 & 76.96 & 85.98 & 86.95 & 94.88 & 85.28 &  & 73.53 & 87.46 & 70.87 & 77.53 & 84.39 & 88.57 & 80.39 \\
TowerInstruct-13B-v0.1 & 78.53 & 89.90 & 77.57 & 87.12 & 87.30 & 95.80 & 86.04 &  & 75.56 & 89.28 & 73.81 & 80.81 & 86.22 & 90.69 & 82.73 \\
DeepTrans-7B\thinkingemoji & / & / & / & / & / & / & / &  & 80.01 & 89.00 & 78.89 & 83.85 & 89.23 & 92.85 & 85.64 \\
GemmaX2-28-9B-v0.1 & 79.40 & 90.63 & 78.71 & 88.60 & 87.85 & 96.33 & 86.92 &  & 77.10 & 90.68 & 75.88 & 83.33 & 87.58 & 92.83 & 84.57 \\
 &  &  &  &  &  &  &  &  &  &  &  &  &  &  &  \\
\textbf{Ours} &  &  &  &  &  &  &  &  &  &  &  &  &  &  &  \\
Qwen2.5-3B & 44.23 & 65.94 & 41.66 & 55.16 & 51.80 & 65.94 & 54.12 &  & 19.81 & 66.64 & 23.99 & 57.69 & 23.31 & 69.16 & 43.43 \\
SSR-Zero-3B & 77.51 & 89.88 & 77.81 & 86.18 & 87.23 & 95.64 & 85.71 &  & 73.88 & 87.14 & 73.57 & 79.40 & 85.43 & 88.66 & 81.35 \\
\\
Qwen2.5-7B & 62.62 & 75.69 & 69.04 & 77.33 & 73.62 & 85.54 & 73.97 &  & 68.25 & 81.63 & 64.28 & 69.48 & 82.00 & 86.07 & 75.29 \\
SSR-Zero-7B\thinkingemoji & 79.29 & 92.04 & 79.04 & 89.19 & 87.97 & 96.70 & 87.37 &  & 79.69 & 91.18 & 79.34 & 85.34 & 89.25 & 93.52 & 86.39 \\
SSR-X-Zero-7B\thinkingemoji & 80.62 & 91.92 & 80.56 & 89.42 & 88.84 & 96.62 & {\underline{88.00}} &  & 81.11 & 91.56 & 79.67 & 86.75 & 90.08 & 93.98 & {\underline{\textbf{87.19}}} \\ \bottomrule
\end{tabular}%
}
\caption{Translation quality measured by COMETKIWI-XXL (KIWI) and XCOMET-XXL (XCM) in English-Chinese directions (EN $\leftrightarrow$ ZH). \textbf{\underline{Bold and underlined}} indicates the best-performing model, \textbf{bold only} the second-best, and \underline{underlined only} the third-best. ``\thinkingemoji'' denotes reasoning models or models operating in thinking mode.}
\label{tab:exp:opensource}
\end{table*}

\begin{table*}[t]
\resizebox{\textwidth}{!}{%
\begin{tabular}{llllllllllllll}
\hline
 & \multicolumn{6}{c}{EN→xx} & & \multicolumn{6}{c}{xx→EN} \\ \cline{2-7} \cline{9-14} 
Models & \multicolumn{2}{c}{EN→GU} & \multicolumn{2}{c}{EN→KK} & \multicolumn{2}{c}{EN→xx} & & \multicolumn{2}{c}{GU→EN} & \multicolumn{2}{c}{KK→EN} & \multicolumn{2}{c}{xx→EN} \\ \cline{2-7} \cline{9-14}
 & KIWI & XCM & KIWI & XCM & KIWI & XCM & & KIWI & XCM & KIWI & XCM & KIWI & XCM \\ \hline
Qwen2.5-7B-Instruct & 19.36 & 23.18 & 18.33 & 10.10 & 18.85 & 16.64 & & 66.95 & 50.96 & 68.11 & 30.80 & 67.53 & 40.88 \\
Qwen2.5-7B & 15.01 & 21.78 & 16.26 & 10.25 & 15.64 & 16.02 & & 48.02 & 44.01 & 58.50 & 26.18 & 53.26 & 35.10 \\
SSR-Zero-7B & 23.47 & 29.30 & 79.05 & 59.06 & \textbf{\underline{51.26}} & \textbf{\underline{44.18}} & & 68.31 & 53.84 & 70.68 & 30.81 & \textbf{\underline{69.50}} & \textbf{\underline{42.33}} \\
\hline
\end{tabular}%
}
\caption{Translation quality across low-resource language pairs, including Gujarati (GU), Kazakh (KK), and aggregated results (xx), measured by COMETKIWI-XXL (KIWI) and XCOMET-XXL (XCM).}
\label{tab:exp:low_resource}
\end{table*}

\subsection{Experimental Setup}
\label{sec:exp:setup}
\paragraph{Dataset} In this paper, we focus on bidirectional translation between English and Chinese, with potential expansion to other language pairs in future work. We use the training dataset released by \citet{feng2025mt}, originally collected from WMT 2017 through WMT 2020 for English-Chinese sentence pairs. Following their preprocessing, sentences shorter than 30 characters were filtered out. Unlike the original bilingual setup, we use these data monolingually, splitting sentence pairs into separate English and Chinese examples to serve as monolingual source sentences for training. The resulting dataset comprises 13,130 monolingual examples (6,565 in English and 6,565 in Chinese).

For testing, we evaluate the translation performance on the English-to-Chinese (EN-ZH) and Chinese-to-English (ZH-EN) benchmarks of WMT23\footnote{https://www2.statmt.org/wmt23/translation-task.html}, WMT24\footnote{https://www2.statmt.org/wmt24/translation-task.html}, and FLORES-200 \cite{costa2022no}.

\paragraph{Metrics} Following the  settings in \citet{rei2024tower}, we adopt two widely used automatic MT-evaluation metrics: the reference-based XCOMET-XXL metric \cite{guerreiro2024xcomet}, and the reference-free COMETKIWI-XXL metric \cite{rei-etal-2023-scaling}, both in their largest available model size.

\paragraph{Baselines}
We compare our models with the following baseline model categories:

\textbf{Closed-source models}, including GPT-4o-20241120 \cite{hurst2024gpt}, Claude-3.5-Sonnet-20240620 \cite{claude_3d5_Sonnet}, and Gemini-1.5-Pro.

\textbf{Open-source general-purpose LLMs}, including the Qwen3 series \cite{yang2025qwen3} (Qwen3-32B, Qwen3-8B), Qwen2.5 series \cite{yang2024qwen2} (Qwen2.5-72B-Instruct, Qwen2.5-32B-Instruct, Qwen2.5-7B-Instruct, Qwen2.5-7B), Qwen's reasoning model QwQ-32B \cite{qwq32b}, and the Gemma2 series \cite{team2024gemma} (Gemma2-27B-it and Gemma2-9B-it).

\textbf{Open-source MT-specific LLMs}, including the Tower series \cite{alves2024tower} (TowerInstruct-7B-v0.2 and TowerInstruct-13B-v0.1), GemmaX2-28-9B-v0.1 \cite{cui2025multilingual}, and DeepTrans-7B \cite{wang2025deep}. 

\paragraph{Implementation Details}

We use Qwen2.5-7B and Qwen2.5-3B as the backbone model and adopt the GRPO algorithm implemented in the verl\footnote{\url{https://github.com/volcengine/verl}} framework. All experiments share the same training settings: a batch size of 128, constant learning rate of 5e-7, rollout number of 16, sampling temperature of 1.0 for generation, and temperature of zero when judging. We set the maximum generation length to 1024 tokens during training. Both KL and entropy coefficients of GRPO are set to zero, as we observed better performance with this configuration.
All models are trained for four epochs using eight GPUs, each providing 148 TFLOPs of computational power when optimizing models with BF16 precision. For training SSR-X-Zero-7B, we add an additional GPU to serve the COMET model. Training SSR-Zero-7B takes about 17 hours, while SSR-X-Zero-7B training takes 42 hours in total.


\subsection{Main Results}

Table \ref{tab:exp:opensource} shows that our SSR-Zero-7B model performs strongly in translation compared to existing open-source models. It achieves an average score of 87.37 in ZH$\rightarrow$EN, outperforming all MT-specific baselines and several larger general-purpose LLMs such as Gemma2-9B-it and Qwen2.5-32B-Instruct. In EN$\rightarrow$ZH, it scores 86.39, surpassing all open-source baselines except Qwen3-32B.

Compared to closed-source models, SSR-Zero-7B slightly lags in ZH$\rightarrow$EN but outperforms GPT-4o in EN$\rightarrow$ZH. It significantly improves upon its backbone model (Qwen2.5-7B) by 18.11\% in ZH$\rightarrow$EN and 14.74\% in EN$\rightarrow$ZH.

Our strongest model, SSR-X-Zero-7B, achieves new SOTA performance among open-source models under 72B parameters, with scores of 88.00 in ZH$\rightarrow$EN and 87.19 in EN$\rightarrow$ZH. It only slightly trails Qwen2.5-72B-Instruct (88.24) in ZH$\rightarrow$EN.

Furthermore, SSR-Zero-3B demonstrates the effectiveness of our approach on smaller base models, improving Qwen2.5-3B's performance by over 58\% in both translation directions.

These results demonstrate the effectiveness and generalizability of leveraging self-generated rewards and external reward models to enhance MT performance.

\section{Analysis}
\label{sec:exp:main_result}

\begin{table*}[]
\resizebox{\textwidth}{!}{%
\begin{tabular}{llllllllllllllll}
\hline
 & \multicolumn{7}{c}{ZH$\rightarrow$EN} &  & \multicolumn{7}{c}{EN$\rightarrow$ZH} \\ \cline{2-8} \cline{10-16} 
Models & \multicolumn{2}{c}{WMT23} & \multicolumn{2}{c}{WMT24} & \multicolumn{2}{c}{Flores200} & \multicolumn{1}{c}{\multirow{2}{*}{Avg.}} &  & \multicolumn{2}{c}{WMT23} & \multicolumn{2}{c}{WMT24} & \multicolumn{2}{c}{Flores200} & \multicolumn{1}{c}{\multirow{2}{*}{Avg.}} \\ \cline{2-7} \cline{10-15}
 & KIWI & XCM & KIWI & XCM & KIWI & XCM & \multicolumn{1}{c}{} &  & KIWI & XCM & KIWI & XCM & KIWI & XCM & \multicolumn{1}{c}{} \\ \hline
Qwen2.5-7B & 62.62 & 75.69 & 69.04 & 77.33 & 73.62 & 85.54 & 73.97 &  & 68.25 & 81.63 & 64.28 & 69.48 & 82.00 & 86.07 & 75.29 \\
\multicolumn{16}{l}{\textbf{w/ External trained MT-evaluation RM:}} \\
{- COMET} & 80.71 & 92.44 & 79.02 & 88.90 & 88.15 & 97.32 & \underline{87.76} &  & 79.80 & 91.95 & 79.54 & 87.11 & 89.30 & 94.54 & {\bf 87.04} \\
{- COMETKIWI} & 79.89 & 91.80 & 81.04 & 89.04 & 89.12 & 96.48 & {\textbf{87.90}} &  & 81.40 & 90.82 & 80.06 & 84.81 & 90.11 & 93.30 & \underline{86.75} \\
\multicolumn{1}{c}{\textbf{}} &  &  &  &  &  &  &  &  &  &  &  &  &  &  &  \\
\multicolumn{16}{l}{\textbf{w/ External LLM-as-a-judge RM (Referenceless):}} \\
{- Qwen2.5-7B} & 78.61 & 91.30 & 78.54 & 87.80 & 87.96 & 96.30 & 86.75 &  & 76.31 & 89.81 & 75.98 & 82.21 & 87.28 & 92.19 & 83.96 \\
{- Qwen2.5-7B-Instruct} & 79.10 & 91.58 & 79.28 & 88.56 & 87.98 & 96.19 & 87.12 &  & 77.03 & 89.73 & 76.60 & 82.16 & 87.87 & 92.07 & 84.24 \\ \\
\multicolumn{16}{l}{\textbf{w/ External LLM-as-a-judge RM (with Reference):}} \\
{- Qwen2.5-7B} & 79.30 & 91.11 & 79.33 & 88.57 & 88.27 & 96.54 & 87.19 &  & 77.90 & 90.00 & 77.69 & 83.43 & 88.38 & 92.63 & 85.01 \\
{- Qwen2.5-7B-Instruct} & 79.10 & 91.58 & 79.28 & 88.56 & 87.98 & 96.19 & 87.12 &  & 77.03 & 89.73 & 76.60 & 82.16 & 87.87 & 92.07 & 84.24 \\
 &  &  &  &  &  &  & {\underline{\textbf{}}} &  &  &  &  &  &  &  &  \\
\multicolumn{16}{l}{\textbf{Ours}} \\
SSR-Zero-7B & 79.29 & 92.04 & 79.04 & 89.19 & 87.97 & 96.70 & 87.37 &  & 79.69 & 91.18 & 79.34 & 85.34 & 89.25 & 93.52 & 86.39 \\
 {- Ablation: w/ ref} & 79.67 & 92.22 & 79.75 & 89.45 & 88.58 & 96.69 & 87.73 &  & 77.91 & 90.62 & 77.63 & 84.15 & \multicolumn{1}{r}{88.25} & 92.96 & 85.25 \\
SSR-X-Zero-7B & 80.62 & 91.92 & 80.56 & 89.42 & 88.84 & 96.62 & {\underline{\textbf{88.00}}} &  & 81.11 & 91.56 & 79.67 & 86.75 & 90.08 & 93.98 & {\underline{\textbf{87.19}}} \\


\hline
\end{tabular}%
}
\caption{\ywj{Translation quality of models trained via RL with different rewarding methods, measured by COMETKIWI-XXL (KIWI) and XCOMET-XXL (XCM) in English-Chinese directions (EN $\leftrightarrow$ ZH). ``Ablation: w/ ref'' denotes a variant of SSR-Zero that includes reference translations in the judge prompt. \textbf{\underline{Bold and underlined}} indicates the best-performing model, \textbf{bold only} the second-best, and \underline{underlined only} the third-best.}}
\label{tab:exp:vs_external}
\end{table*}
Although SSR and its combination with external reward models (RMs) effectively enhance MT performance, two research questions (RQs) remain unclear: 
1) \textit{Can SSR generalize to other languages, especially low-resource ones?}
2) \textit{How does self-rewarding compare with widely used external RMs?} 3) \textit{How does the inclusion of reference data in RMs affect the final translation performance?} To clarify these points, we conducted a detailed analysis, presented below.

\subsection{RQ1: SSR for Low-Resource Languages}
To evaluate the generalizability of SSR for low-resource languages, we selected Kazakh (KK) and Gujarati (GU) from the WMT19 dataset. We sampled 3k monolingual sentences per language from the WMT19 training set for training and evaluated the models using the entire test set (2k per language).

\paragraph{Result}
Table \ref{tab:exp:low_resource} shows significant performance gains from SSR training based on the Qwen2.5-7B model. For EN $\rightarrow$ xx translation, COMETKIWI scores improved substantially from 15.64 to 51.26 (+227\%), and XCOMET scores rose from 16.02 to 44.18 (+175\%). \ywj{We note that these large relative gains partly reflect the low absolute baselines typical of low-resource language pairs.} For xx$\rightarrow$EN translation, COMETKIWI scores increased from 53.26 to 69.50 (+30.49\%), while XCOMET scores improved from 35.10 to 42.33 (+20.59\%). SSR-Zero-7B significantly outperformed Qwen2.5-7B-Instruct, indicating that our approach effectively generalizes and is particularly beneficial in low-resource scenarios.

\subsection{RQ2: SSR vs. External Reward Models}


 Specifically, we compare our method with two categories of external frozen RMs: 1) MT-evaluation trained RMs, including COMET\footnote{\url{https://huggingface.co/Unbabel/wmt22-comet-da}} and COMETKIWI\footnote{\url{https://huggingface.co/Unbabel/wmt22-cometkiwi-da}}, and 2) LLM-based judge RMs, including Qwen2.5-7B and Qwen2.5-7B-Instruct, using the same judge prompts employed by SSR.


\paragraph{Results}

The evaluation results are summarized in Table \ref{tab:exp:vs_external}. As expected, models trained with specialized MT-evaluation RMs (i.e., COMET or COMETKIWI) outperform SSR-Zero-7B -- which relies solely on intrinsic judgments from the training model  -- in average EN $\rightarrow$ ZH translation scores. Additionally, these specialized RMs also outperform all methods using external LLM-as-a-judge approaches based on the 7B-sized Qwen2.5 model. This indicates that dedicated RMs trained on large annotated datasets possess stronger MT evaluation capabilities compared to general-purpose LLMs such as Qwen2.5-7B(-Instruct).
Nevertheless, the SSR mechanism provides complementary benefits. This is evidenced by SSR-X-Zero-7B, which integrates self-rewarding with COMET supervision, still achieves the highest scores in both translation directions.

Furthermore, SSR-Zero-7B substantially outperforms models with the same backbone trained using external LLM judges of the same size. This indicates that, during SSR training, improvements in translation capability may simultaneously enhance a model's judgment ability.

\ywj{
\paragraph{Why does self-rewarding work?} An intuitive concern is that a model with weak translation capabilities should produce unreliable evaluation scores. However, we argue that the base model is not inherently weak in capability, but rather unaligned. As shown in Table~\ref{tab:exp:fewshot}, the backbone already demonstrates strong translation potential under few-shot prompting, confirming that the capability exists latently. SSR effectively unlocks this potential by using the model's internal judge to provide informative reward signals for RL training. Additionally, GRPO computes advantages via group-level reward normalization, making training depend on relative rankings within a batch rather than absolute scores, which provides robustness to noisy self-judgments. The convergence of SSR-Zero-3B (Table~\ref{tab:exp:opensource}) -- a smaller model with weaker judging capability -- further supports this. Nonetheless, a quantitative diagnostic of judge--external metric disagreement remains a valuable direction for future work.
\label{sec:exp:why_ssr_works}
}




\subsection{RQ3: Reference vs. Referenceless Rewarding}
We further examine the influence of reference translations on reward signals and their subsequent impact on MT performance. Specifically, we introduce a variant of SSR-Zero that includes a reference translation in the judge prompt\ywj{, denoted as ``Ablation: w/ ref'' in Table~\ref{tab:exp:vs_external}}. The reference translation is obtained using the original target sentence from the training dataset. We use the same setting for LLM-as-a-judge baselines.


\paragraph{Results}
As shown in Table \ref{tab:exp:vs_external}, the trained reference-based RM (COMET) and referenceless RM (COMETKIWI) yield similar results. For LLM-based external judges, explicitly providing reference translations typically leads to slightly higher performance compared to the reference-less setting. 
In self-reward training, the use of reference translations marginally improves performance in ZH $\rightarrow$ EN translation (from 87.37 to 87.73, +0.4\%), but lowers the results for EN $\rightarrow$ ZH translation (from 86.39 to 85.25, -1.3\%). In general, introducing reference translations to different reward methods does not consistently improve the model's performance, except when using external LLMs as judges. \ywj{In particular, external references do not provide significant gains for SSR.}

\ywj{
\subsection{Additional Studies}

\paragraph{Multilingual Generalization} We evaluate SSR-Zero-7B (trained only on $\sim$13K EN$\leftrightarrow$ZH sentences) on FLORES-200 covering 33 languages using XCOMET-XXL (Table~\ref{tab:exp:multilingual}). Despite never seeing other language pairs during training, SSR-Zero-7B improves translation quality across diverse directions, including X$\rightarrow$Y pairs not involving English or Chinese (+3.3 on average). We attribute this to shared cross-lingual representations in the pretrained backbone and the universal nature of the quality criteria learned through self-judging. Representative per-pair results are in Appendix~\ref{sec:appendix:flores}. That said, training remains centered on EN$\leftrightarrow$ZH, and generalization claims should be interpreted accordingly.

\paragraph{Human Evaluation} We conduct a segment-level Direct Assessment (DA) on WMT23 EN$\leftrightarrow$ZH (200 samples per direction, two bilingual annotators). Scores were z-normalized per annotator following WMT protocols. As shown in Table~\ref{tab:exp:human_eval}, SSR-Zero-7B achieves higher mean DA scores than Qwen2.5-7B-Instruct in both directions, confirming that automatic metric gains correspond to real improvements in perceived quality.

\paragraph{Comparison with Few-Shot Prompting} We compare SSR-Zero-7B with few-shot prompting baselines (0/1/5/10-shot) using the same backbone (Table~\ref{tab:exp:fewshot}). Few-shot prompting improves the backbone but saturates after one shot, reaching a level comparable to Qwen2.5-7B-Instruct. SSR-Zero-7B substantially outperforms all variants without any in-context examples, demonstrating that the gains from self-rewarding RL training cannot be replicated by simply providing parallel examples at inference time.

\begin{table}[t]
\centering
\small
\resizebox{\columnwidth}{!}{
\begin{tabular}{lccc}
\hline
Direction & Qwen2.5-7B-Inst. & SSR-Zero & $\Delta$ \\
\hline
ZH$\rightarrow$X & 50.7 & \textbf{52.2} & +1.5 \\
X$\rightarrow$ZH & 69.9 & \textbf{71.1} & +1.2 \\
EN$\rightarrow$X & 50.7 & \textbf{53.6} & +2.9 \\
X$\rightarrow$EN & \textbf{79.0} & 78.9 & -0.1 \\
X$\rightarrow$Y (non-EN/ZH) & 38.9 & \textbf{42.2} & +3.3 \\
\hline
All & 41.7 & \textbf{44.9} & +3.2 \\
\hline
\end{tabular}}
\caption{Multilingual evaluation on FLORES-200 (33 languages, XCOMET-XXL). SSR-Zero-7B is trained only on EN$\leftrightarrow$ZH.}
\label{tab:exp:multilingual}
\end{table}

\begin{table}[t]
\centering
\small
\resizebox{\columnwidth}{!}{
\begin{tabular}{llccc}
\hline
Direction & Model & Mean DA & z-score & Std \\
\hline
\multirow{2}{*}{EN$\rightarrow$ZH} & SSR-Zero-7B & \textbf{85.65} & +0.073 & $\pm$0.90 \\
 & Qwen2.5-7B-Inst. & 82.90 & -0.073 & $\pm$1.00 \\
\hline
\multirow{2}{*}{ZH$\rightarrow$EN} & SSR-Zero-7B & \textbf{91.05} & +0.050 & $\pm$0.80 \\
 & Qwen2.5-7B-Inst. & 89.69 & -0.050 & $\pm$0.97 \\
\hline
\end{tabular}}
\caption{Human Direct Assessment on WMT23 EN$\leftrightarrow$ZH.}
\label{tab:exp:human_eval}
\end{table}

\begin{table}[t]
\centering
\small
\begin{tabular}{lcc}
\hline
Model & ZH$\rightarrow$EN & EN$\rightarrow$ZH \\
\hline
Qwen2.5-7B (0-shot) & 79.52 & 79.06 \\
Qwen2.5-7B (1-shot) & 89.39 & 84.37 \\
Qwen2.5-7B (5-shot) & 89.29 & 84.82 \\
Qwen2.5-7B (10-shot) & 89.36 & 84.75 \\
Qwen2.5-7B-Instruct & 90.19 & 86.31 \\
\hline
SSR-Zero-7B (0-shot) & \textbf{92.64} & \textbf{90.01} \\
\hline
\end{tabular}
\caption{Average XCOMET-XXL scores for few-shot baselines vs.\ SSR-Zero-7B.}
\label{tab:exp:fewshot}
\end{table}
}

\section{Conclusion}
\label{sec:conclusion}

In this work, we propose \emoji\textbf{SSR}, a simple yet effective reinforcement learning approach for machine translation. SSR does not rely on external reward models (RMs) or reference data; instead, it leverages the actor model itself as a judge to generate reward signals and optimize its performance through GRPO training. Initialized from an uninstructed Qwen2.5-7B backbone, our SSR-Zero-7B model outperforms many open-source MT-specific LLMs, such as TowerInstruct-13B, as well as larger general-purpose LLMs like Qwen2.5-32B-Instruct across multiple English $\leftrightarrow$ Chinese translation benchmarks.

Our analysis shows that SSR is more effective than using same-size external LLM-as-a-judge models. In addition to high-resource settings, SSR demonstrates strong generalization to low-resource language pairs, yielding substantial improvements when trained with limited monolingual data. Although SSR alone slightly underperforms dedicated RMs (e.g., COMET and COMETKIWI) trained on large-scale annotated MT-evaluation data, combining SSR with these RMs yields consistent additional improvements. Our best-performing model, SSR-X-Zero-7B, integrates SSR with COMET and achieves competitive performance relative to existing open-source and closed-source systems on English $\leftrightarrow$ Chinese translation benchmarks.

These findings provide insight into reward selection for MT via reinforcement learning and highlight that strong pre-trained LLMs inherently possess reliable MT evaluation capabilities that can be exploited to improve translation quality. Overall, our work demonstrates the potential of self-reward-based RL approaches to reduce dependence on costly external supervision from humans or trained reward models, particularly in low-resource scenarios.

\section*{Limitations}

Our work demonstrates the effectiveness and, to some extent, the generalizability of self-reward training for machine translation. However, the applicability of this approach across different model architectures remains unexplored. Prior work has shown that R1-Zero-like training can exhibit varying effectiveness across model families \cite{gandhi2025cognitive}, and it therefore remains unclear whether SSR can consistently incentivize strong MT capabilities in architectures such as Llama.

\ywj{Although our few-shot comparison (Table~\ref{tab:exp:fewshot}) shows that providing in-context examples to the actor does not match SSR's gains, the effect of alternative prompting strategies for the judge -- such as Chain-of-Thought (CoT) or few-shot prompting -- remains unexplored.} That said, recent work by \citet{qian2024large} suggests that CoT and 5-shot prompting do not outperform zero-shot prompting for MT evaluation when using 7B-scale models with similar evaluation prompts.

In addition, using the model itself as a judge may introduce a potential risk of reinforcing incorrect self-judgments or exhibiting reward bias. \ywj{As discussed in Section~\ref{sec:exp:why_ssr_works}, our evidence suggests that GRPO's group-level normalization and the backbone's latent capabilities provide sufficient robustness, and the qualitative analysis in Appendix~\ref{sec:appendix:qualitative} shows that the judge can identify major errors despite occasional misjudgments. Nevertheless, a deeper theoretical and empirical characterization of failure modes in self-rewarding remains necessary, particularly for weaker backbone models.}

Another limitation is that our evaluation relies \ywj{heavily} on automatic metrics rather than direct human assessment. We use XCOMET-XXL and COMETKIWI-XXL, which are recommended by the WMT community\footnote{\url{https://www2.statmt.org/wmt25/translation-task.html}} and have been shown to correlate strongly with human judgments. \ywj{Although our small-scale human evaluation (Table~\ref{tab:exp:human_eval}) confirms that metric gains correspond to real quality improvements, a larger-scale human assessment covering more systems and language pairs remains an important direction for future work.}

\ywj{Furthermore, our training is centered on EN$\leftrightarrow$ZH. While Section 5.4 demonstrates encouraging cross-lingual transfer to 33 unseen languages, systematic evaluation across more diverse training language pairs is needed to fully validate the generalizability of SSR.}

Finally, recent studies \cite{liu2025inference} indicate that LLM-as-a-judge frameworks may benefit from test-time scaling techniques such as voting. We leave the exploration of such techniques within the context of SSR-based training to future work.

\ywj{

}
\bibliography{anthology,custom}

\appendix

\section{Appendix}

\subsection{Training Dynamics of SSR}

\begin{figure}[h]
  \includegraphics[width=\columnwidth]{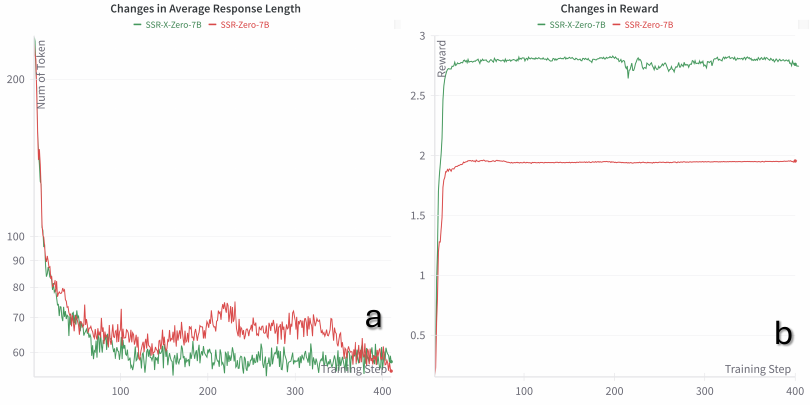}
  \caption{Changes in average response length (a) and training rewards (b) of SSR/SSR-X-Zero-7B during GRPO training.}
  \label{fig:resp_len}
\end{figure}

We also report how the response length  and test set performance evolve during SSR/SSR-X-Zero-7B training.  As shown in Fig. \ref{fig:resp_len}, we did not observe the increase in output length typical of R1-like training in mathematics \cite{guo2025deepseek}, nor the curve seen in \citet{feng2025mt} which first decreases and then increases. As training progressed, the model quickly reduced the output length from about 200 to 60-70 tokens and did not generate meaningful CoTs. 
A typical CoT before translation was ``\texttt{<think>} \textit{I need to translate this sentence from \{src\_lang\} to \{tgt\_lang\}.}\texttt{</think>}''. 

\begin{figure}[h]
  \includegraphics[width=\columnwidth]{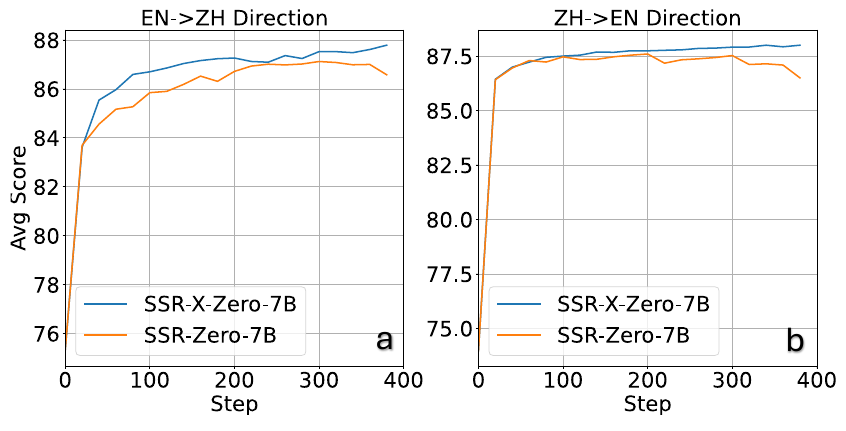}
  \caption{Changes in translation quality during training, measured by the average scores of COMETKIWI-XXL and XCOMET-XXL on the EN $\rightarrow$ ZH (a) and ZH $\rightarrow$ EN (b) benchmarks.}
  \label{fig:score_dyn}
\end{figure}
Despite this, we observed an increase trend in performance in the test set as training progressed, as shown in Figure \ref{fig:score_dyn}. We also noticed that the performance of SSR-Zero-7B for EN $\rightarrow$ ZH saturates after approximately 3 epochs (around 300 steps) and decreases afterward, while its ZH $\rightarrow$ EN performance converges earlier, at roughly 200 steps. In contrast, SSR-Zero-X-7B demonstrates better stability and continuous improvement during training. Upon inspection, we found that SSR-Zero-7B began enclosing translated outputs with extraneous quotation marks (i.e., \texttt{<answer>}``translated text''\texttt{</answer>}) after 300 steps, which our regular expression could not filter out during evaluation. This formatting issue led automated metrics XCOMET-XXL and COMETKIWI-XXL to produce lower evaluation scores. This issue was not observed during the SSR-Zero-X-7B's training. We leave further exploration in maintaining consistent output formatting of SSR training for future work.

\ywj{
\subsection{FLORES-200 Per-Pair Results}
\label{sec:appendix:flores}

Table~\ref{tab:appendix:flores_pairs} shows representative per-pair results from our FLORES-200 evaluation (33 languages) using XCOMET-XXL. SSR-Zero-7B is trained only on EN$\leftrightarrow$ZH data. \textbf{Bold} indicates the better score between Qwen2.5-7B-Instruct and SSR-Zero-7B.

\begin{table}[h]
\centering
\small
\begin{tabular}{lcc}
\hline
Direction & Qwen2.5-7B-Inst. & SSR-Zero \\
\hline
EN$\rightarrow$DE & 90.4 & \textbf{93.1} \\
EN$\rightarrow$ES & 88.3 & \textbf{93.4} \\
EN$\rightarrow$FR & 82.7 & \textbf{89.6} \\
EN$\rightarrow$JA & 67.5 & \textbf{78.5} \\
EN$\rightarrow$KO & 50.8 & \textbf{61.3} \\
EN$\rightarrow$VI & 73.2 & \textbf{81.2} \\
EN$\rightarrow$RU & 72.7 & \textbf{79.4} \\
EN$\rightarrow$PT & 88.1 & \textbf{92.0} \\
\hline
DE$\rightarrow$EN & 94.5 & \textbf{95.4} \\
JA$\rightarrow$EN & 90.0 & \textbf{91.4} \\
KO$\rightarrow$EN & 88.7 & \textbf{90.9} \\
ID$\rightarrow$EN & 92.8 & \textbf{93.4} \\
\hline
ZH$\rightarrow$DE & 80.9 & \textbf{88.1} \\
ZH$\rightarrow$ES & 82.6 & \textbf{90.4} \\
ZH$\rightarrow$KO & 56.8 & \textbf{68.4} \\
ZH$\rightarrow$RU & 73.5 & \textbf{83.1} \\
ZH$\rightarrow$MS & 68.0 & \textbf{76.9} \\
\hline
RU$\rightarrow$ZH & 87.0 & \textbf{88.9} \\
ES$\rightarrow$ZH & 88.2 & \textbf{91.2} \\
IT$\rightarrow$ZH & 83.1 & \textbf{86.6} \\
PT$\rightarrow$ZH & 84.7 & \textbf{89.1} \\
\hline
\end{tabular}
\caption{Representative per-pair XCOMET-XXL scores on FLORES-200. SSR-Zero-7B is trained only on EN$\leftrightarrow$ZH data.}
\label{tab:appendix:flores_pairs}
\end{table}
}

\ywj{
\subsection{Qualitative Analysis of Self-Judging}
\label{sec:appendix:qualitative}

We present qualitative examples illustrating (a) translation improvements from backbone to SSR-Zero, (b) cases where the self-judge produces accurate assessments, and (c) cases where it misjudges.

\paragraph{Translation Improvements}

\textbf{Case 1 (Subject misinterpretation):}
\begin{itemize}[nosep,leftmargin=*]
\item \textbf{Source:} ``\begin{CJK}{UTF8}{gbsn}我听说她被艾尔文王后附身了，所以总是去那个房间。\end{CJK}''
\item \textbf{Qwen2.5-7B:} ``I heard she was possessed by Queen Elvina, so \textbf{I} always go to that room.'' (wrong subject)
\item \textbf{SSR-Zero-7B:} ``I heard that she was possessed by Queen Elvin, which is why \textbf{she} always went to that room.'' \checkmark
\end{itemize}

\textbf{Case 2 (Literal translation error):}
\begin{itemize}[nosep,leftmargin=*]
\item \textbf{Source:} ``\begin{CJK}{UTF8}{gbsn}创新高了。\end{CJK}''
\item \textbf{Qwen2.5-7B:} ``\textbf{Innovations peaked.}'' (confuses ``\begin{CJK}{UTF8}{gbsn}创新高\end{CJK}/hit a new high'' with ``\begin{CJK}{UTF8}{gbsn}创新\end{CJK}/innovation'')
\item \textbf{SSR-Zero-7B:} ``\textbf{It set a new record.}'' \checkmark
\end{itemize}

\paragraph{Accurate Self-Judgments}

\textbf{Case 1 (Detecting name mistranslation):}
\begin{itemize}[nosep,leftmargin=*]
\item \textbf{Source:} ``In the fourth set, Zhu Yuling made a good start and was 4-2 ahead.''
\item \textbf{Actor:} ``\begin{CJK}{UTF8}{gbsn}第四局, \textbf{荣钰婷}开局不错, 导致了4-2的领先。\end{CJK}''
\item \textbf{Judge:} ``The Chinese translation mentions `\begin{CJK}{UTF8}{gbsn}荣钰婷\end{CJK}' instead of `Zhu Yuling' and adds `\begin{CJK}{UTF8}{gbsn}导致了\end{CJK} (caused)' not in the original.'' $\rightarrow$ Score: 70
\end{itemize}

\textbf{Case 2 (Detecting repetition):}
\begin{itemize}[nosep,leftmargin=*]
\item \textbf{Source:} ``\begin{CJK}{UTF8}{gbsn}军训日晒时间较长，防晒霜每隔两小时擦一次才能有持久效果，千万不要偷懒。\end{CJK}''
\item \textbf{Actor:} ``...never slack off \textbf{slacking off}.''
\item \textbf{Judge:} ``It contains repetitive content such as `slacking off' and `never slack off,' and uses informal language.'' $\rightarrow$ Score: 70
\end{itemize}

\paragraph{Inaccurate Self-Judgments}

\textbf{Case 1 (Overlooking fluency):}
\begin{itemize}[nosep,leftmargin=*]
\item \textbf{Source:} ``ABC was second with 3.26 million viewers.''
\item \textbf{Actor:} ``ABC\begin{CJK}{UTF8}{gbsn}获得了326万观众的\textbf{第二名}。\end{CJK}'' (awkward; more natural: ``ABC\begin{CJK}{UTF8}{gbsn}以326万观众位居第二\end{CJK}'')
\item \textbf{Judge:} ``The grammar and meaning are preserved.'' $\rightarrow$ Score: 100 (expected $\sim$85)
\end{itemize}

\textbf{Case 2 (Hallucinating an error):}
\begin{itemize}[nosep,leftmargin=*]
\item \textbf{Source:} ``\begin{CJK}{UTF8}{gbsn}美国乘客Bill Jaynes称，这架飞机飞得很低。\end{CJK}''
\item \textbf{Actor:} ``\textbf{American} passenger Bill Jaynes said that the plane was flying very low.''
\item \textbf{Judge:} ``It is missing the name of the country where the passengers are from.'' $\rightarrow$ Score: 80 (expected 100)
\end{itemize}

These examples show that the self-judge can identify major errors such as name mistranslation and repetition, yet may overlook fluency issues or hallucinate errors. Despite these imperfections, the self-judge provides a sufficiently robust reward signal for RL training, as evidenced by consistent improvements on both automatic metrics and human evaluation.
}

\end{document}